# Learning Undirected Graphical Models with Structure Penalty

**Shilin Ding** sding@stat.wisc.edu
*Department of Statistics*
*University of Wisconsin-Madison*
*Madison, WI 53706, USA*

April 26, 2011

## Abstract

In undirected graphical models, learning the graph structure and learning the functions that relate the predictive variables (features) to the responses given the structure are two topics that have been widely investigated in machine learning and statistics. Learning graphical models in two stages will have problems because graph structure may change after considering the features. The main contribution of this paper is the proposed method that learns the graph structure and functions on the graph at the same time. General graphical models with binary outcomes conditioned on predictive variables are proved to be equivalent to multivariate Bernoulli model. The reparameterization of the potential functions in graphical model by conditional log odds ratios in multivariate Bernoulli model offers advantage in the representation of the conditional independence structure in the model. Additionally, we impose a structure penalty on groups of conditional log odds ratios to learn the graph structure. These groups of functions are designed with overlaps to enforce hierarchical function selection. In this way, we are able to shrink higher order interactions to obtain a sparse graph structure. Simulation studies show that the method is able to recover the graph structure. The analysis of county data from Census Bureau gives interesting relations between unemployment rate, crime and others discovered by the model.

## 1. Introduction

In undirected graphical models (Markov Random Fields), a graph $G$ is defined as $G = (V, E)$, where $V = \{1, \cdots, K\}$ is the set of nodes and $E \subset V \times V$ is the set of links between the nodes. In fact, $V$ is associated to a set of multivariate response variables $Y_1, \cdots, Y_K$, and $E$ specifies the conditional independence structure among them. For example, a link between two nodes $i, j$ indicates a pairwise interaction $f^{ij}$. , and a clique between three nodes $i, j, k$ indicates a third order interaction $f^{ijk}$. These functions formulate the effects of predicative variables (features), $X$, on the responses and their interactions.

Graphical models have been used in many applications. Conditional Random Fields (CRF) (Lafferty et al., 2001) and their extensions, e.g. dynamic CRF (Sutton et al., 2007), are well known in Natural Language Processing community. The CRFs achieve great success in sequentially structured text, by modeling the interaction of labels $(Y)$ on the nodes conditioned flexibly on the features $(X)$. There are also numerous applications of graphical models to computer vision (Szeliski et al., 2007; Honorio and Samaras, 2010). Ising model is another classical example that draws great interest in image processing (Williams et al., 2004) and also has been recently applied to social networks (Banerjee et al., 2008).

The intuition of utilizing a graph structure is that some responses are related while others are not. However, in many cases, the graph is pre-determined by domain knowledge. For example, Duan et al. (2008) proposed a collective model for labeling music signals with fully connected graph, which they called collective conditional random fields. They have 10 semantic categories such as genre (blues, rap, ... ), instrument (guitar, piano, ... ), production (studio, live), rhythm(strong, weak, middle), and etc. It is possible that some links are not necessary: e.g. production and instrument. Estimating the parameters on these relations will lead to over-fitting. Therefore, graph structure learning is an important aspect of relation discovery in multivariate response applications and multi-task learning.

Many papers have focused on the graphical model selection issue. Meinshausen and Buhlmann (2006) and Peng et al. (2009) studied sparse covariance estimation of Gaussian outcomes (Speed and Kiiveri, 1986) without input features. The covariance matrix determines the dependence structure in the Gaussian distribution and its sparsity specifies the linkage in Gaussian Markov Random Fields. This is not the case for non-elliptical distribution, such as the distribution of discrete random variables. Ravikumar et al. (2010) focused on graph structure selection of Ising model based on $l_1$-regularized logistic regression. It gave sufficient conditions for consistently estimating the neighborhood of the nodes, without input features. However, two marginally independent response variables may become dependent after conditioning on $X$. So, ignoring the predicative variables may lead to inconsistent estimation of the graph structure. To the best of our knowledge, there is no previous work addressed the issue of learning the graph structure and the functions associated with the graph at the same time.

In this paper, our first contribution is the proof of the equivalence between the general graphical model with bivariate outcomes and multivariate Bernoulli (MVB) model. The functions that represent the effects of predicative variables on responses and their interactions (at all levels) can be formulated in MVB model, which is endowed with the advantage of interpreting the graph structure. It follows from the sparsity of links in the graphical models that some functions are constant zero, which means certain responses are conditionally independent. Therefore, we impose the structure penalty on groups of functions with overlaps to obtain sparse estimation of the graph structure. These groups are designed to enforce the sparsity on the functions and shrink higher order interactions so that they only appear after their lower components have entered the model.

The paper is organized as follows: Section 2 introduces graphical models and their relation with multivariate Bernoulli model. Section 3 and 4 discusses the model for learning the graph structure and the functions on the graph through structure penalty. The experiments are shown and discussed in Section 5. Section 6 gives the conclusion and future work.

## 2. Conditional Independence in Graphical Models

The notations in this paper are summarized in Table 1.

Table 1: Notations

| Symbol | Description |
|---|---|
| $\|\cdot\|$ | $L_2$ norm |
| $n$ | Sample size |
| $p$ | Number of covariates |
| $K$ | Number of Response/Output |
| $Y$ | $K$ dimensional Response/Output |
| $\Omega$ | Set of $\{1, 2, \ldots, K\}$ |
| $\Psi_K$ | Power set of $\Omega$ except the empty set |
| $m$ | The highest level of interaction considered in the model, in this paper $m = K$ |
| $q$ | Number of $f^\omega$ in the assumption, in this paper $q = 2^K - 1$ |
| $\omega, \kappa, v$ | Element of $\Psi_K$ used for indexing |
| $y^\omega(i)$ | $y^\omega(i) = \prod_{k \in \omega} y_k(i)$ |
| $\mathcal{Y}(i)$ | Augmented responses $(y^1(i), \ldots, y^\Omega(i))$ |
| $c$ | Model parameters |
| $\tilde{p}$ | $(p+1) \cdot q$, dimension of $c$ |
| $T_v$ | $T_v = \{\omega \| v \subset \omega\}$ is the subgraph rooted at $v$ containing all its descendants |
| $f^{T_v}$ | $f^{T_v} = (f^\omega), \omega \in T_v$ |
| $J(f^{T_v})$ | Penalty on $f^{T_v}$ |
| $p_v$ | Weight for penalty on structure $T_v$ |
| $s_v, r_v$ | Subgradient of $\lambda J(f^{T_v})$ of the $v$th group |
| $S^\omega(y; x)$ | $S^\omega(y; x) = \sum_{\kappa \in T^\omega} y^\kappa f^\kappa$ |

In graphical models, $G = (V, E)$, the distribution of multivariate discrete random variables $Y_1, \ldots, Y_K$ is:

$$p(Y_1 = y_1, \ldots, Y_K = y_K | X) = \frac{1}{Z(X)} \prod_{C \in \mathcal{C}} \Phi_C(y_C; X) \qquad (1)$$

where $Z$ is the normalization factor. The distribution is factorized according to the cliques in the graph. A clique $C \subset \Omega = \{1, \ldots, K\}$ is the set of nodes of a fully connected subgraph. $\Phi_C(y_C; X)$ is the potential function on $C$. It depends on $y_C = \{y_i \mid i \in C\}$ and the predicative variables $X$ which are shared across all response variables. One example of application is to model the relations of multiple clinical responses (hypertension, diabetes, etc.) and how they are affected by the person's genetic variables and environmental variables (smoking, income, etc.).

For the purpose of efficient computation, $\mathcal{C}$ is usually the set of all maximal cliques of the graph. The maximal clique is a clique that is not properly contained in any other clique in the graph. Different representations with $\mathcal{C}$ as the set of non-maximal cliques can be converted to maximal clique representation by redefinition of the potential functions (Wainwright and Jordan, 2008). Furthermore, $\mathcal{C}$ does not have to reflect the graph structure, as long as it is sufficient. For example, the most general choice for any given graphical model is $\mathcal{C} = \{\Omega\}$. The conditional independence between the response variables is implicitly formulated by the restrictions on the potential functions. See Theorem 2.2 and Example 2.1 for details.

The Markov property states that any two nodes not in a clique are conditionally independent given other nodes. For example, $Y_s, Y_t$ are conditionally independent given all other variables that block the path from $Y_s$ to $Y_t$. Therefore, $\mathcal{C}$ as the set of maximal cliques factorizes the graph and specifies the conditional independence in the model. In Figure 1(a), we have 2 cliques $\{1, 2, 3\}$ and $\{3, 4\}$. In this case, $\{Y_2, Y_4\}$ are conditionally independent given $Y_3$, so are $\{Y_1, Y_4\}$.

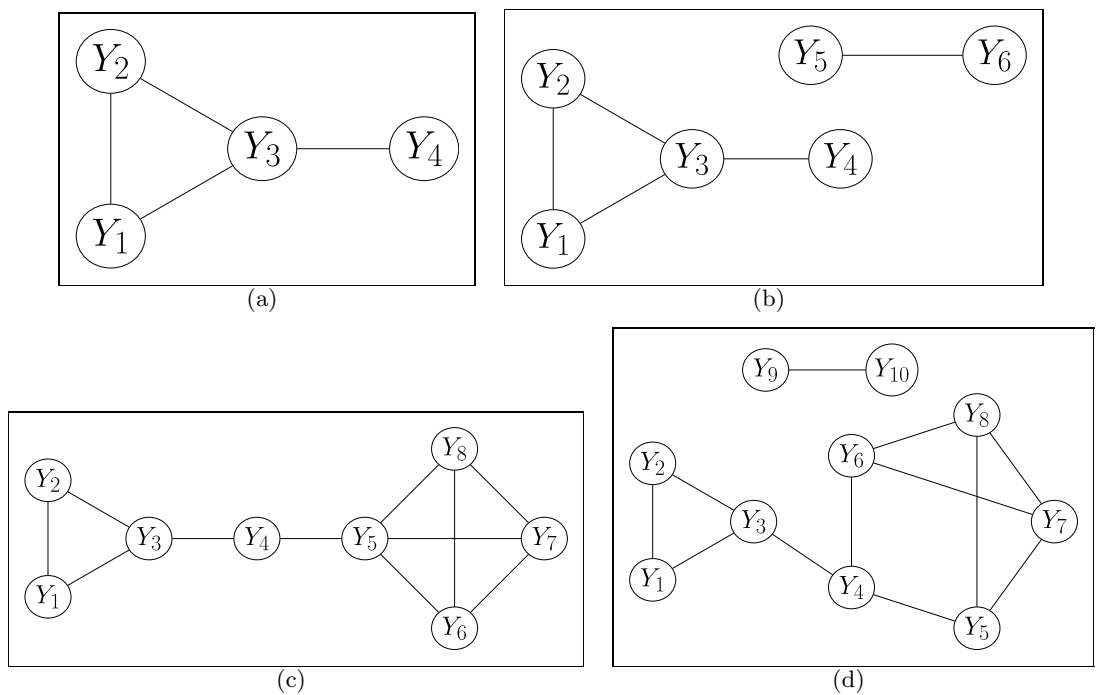

Figure 1: Graphical Model Examples. (a) Model 1: The triangle clique $\{Y_1, Y_2, Y_3\}$ indicate a third order interaction. Additionally, there is a pairwise interaction between $Y_3$ and $Y_4$. $Y_4$ is conditionally independent with $Y_1$ or $Y_2$ given $Y_3$. (b) Model 2: $Y_5, Y_6$ are another set of interacted random variables which are independent of other 4 nodes. (c) Model 3: $Y_5, Y_6, Y_7, Y_8$ form a 4-node clique that connected to $Y_1, Y_2, Y3$ through $Y_4$. (d) Model 4: $Y_4, \cdots, Y_8$ form a strongly connected component, and $Y_9, Y_{10}$ are independent of other nodes

Given the graph structure, the potential functions are convenient to characterize the distribution on the graph. However, if the graph is unknown in advance, estimating the potential functions does not give direct inference of the graph structure, because there are different representations with different choices of cliques in the same graph as mentioned above. Even if assuming the most general case where $\mathcal{C} = \{\Omega\}$, the conditional independence between the nodes cannot be represented by $\Phi_C$ in a simple form (see Example 2.1), which makes the learning of graph structure difficult. The multivariate Bernoulli (MVB) model can represent a graphical model whose nodes are Bernoulli random variables, i.e. $Y_i = 0$ or 1. And the parameterization in MVB model is suitable for learning the graph structure. We will show later that it is equivalent to GM (1) with binary outcomes. The distribution of MVB model is:

$$p(Y_1 = y_1, \ldots, Y_K = y_k | X)$$
$$= \exp\{y_1 f^1 + \cdots + y_K f^K + \cdots + y_1 y_2 f^{1,2} + \cdots + y_1 \ldots y_K f^{1,\ldots,K} - b(f)\}$$
$$= \exp\{\sum_{\omega=1}^{2^K-1} y^\omega f^\omega - b(f)\} \quad (2)$$

Here, we use the following notations. Let $\overline{\Psi}_K$ be the power set of $\Omega = \{1, \ldots, K\}$. Counting the empty set, there are $2^K$ elements in $\overline{\Psi}_K$. Where convenient in what follows, we will relabel these elements from 0 to $2^K - 1$, e.g. for $K = 3$, we will use $f^{1,2,3}$ and $f^7$ interchangeably without further specification. Because there are $2^K - 1$ free parameters in (2), denote $\Psi_K = \overline{\Psi}_K - \{\emptyset\}$ for simple notation. Let $\omega$ denotes a set in $\Psi_K$, define $\mathcal{Y} = (y^1, \cdots, y^\omega, \cdots, y^\Omega)$ be the augmented response with

$$y^\omega = \prod_{i \in \omega} y_i \quad (3)$$

And $f = (f^1, \ldots, f^\omega, \ldots, f^\Omega)$ be the vector of natural parameters, where $f^\omega(x)$ is the conditional log odds ratio (Gao et al., 2001) to be estimated

$$f^\omega = \log OR(Y_i, i \in \omega | Y_j = 0, j \notin \omega; X) \quad (4)$$

Here, the odds ratios are calculated recursively

$$OR(Y_i) = \frac{P(Y_i = 1)}{1 - P(Y_i = 1)}, \quad (5)$$

$$OR(Y_i, i \in \omega \cup \{k\}) = \frac{OR(Y_i, i \in \omega | Y_k = 1)}{OR(Y_i, i \in \omega | Y_k = 0)}, \text{ suppose } k \notin \omega \quad (6)$$

Later, we will call $f^1, \cdots, f^K$ main effects, and $f^{1,2}, \cdots, f^{1,\cdots,K}$ measures the interactions between the response variables. We are interested in the sparse estimation of $f^\omega$, which is in a Reproducing Kernel Hilbert Space (RKHS) $H^\omega$ with kernel $K^\omega$ (Wahba, 1990). Denote:

$$S^\omega(y; x) = \sum_{\kappa \subset \omega} y^\kappa f^\kappa(x); \quad S^\omega(x) = \sum_{\kappa \subset \omega} f^\kappa(x); \quad (7)$$

Then the normalization factor is:

$$\exp(b(f(x))) = 1 + \sum_{\omega \in \Psi_K} \exp(S^\omega(x)) \quad (8)$$

And we have the following lemma:

**Lemma 2.1** *In multivariate Bernoulli model, define the odd-even partition of the power set of $\omega$ as: $\Psi^\omega_{odd} = \{\kappa \subset \omega \mid |\kappa| = |\omega| - k, \text{where } k \text{ is odd}\}$, and $\Psi^\omega_{even}\{\kappa \subset \omega \mid |\kappa| = |\omega| - k, \text{where } k \text{ is even}\}$. Note $|\Psi^\omega_{odd}| = |\Psi^\omega_{even}| = 2^{|\omega|-1}$, the natural parameters have the following property:*

$$f^\omega = \log \frac{\prod_{\kappa \in \Psi^\omega_{even}} p(Y_i = 1, i \in \kappa, \text{and } Y_j = 0, j \in \Omega - \kappa | X)}{\prod_{\kappa \in \Psi^\omega_{odd}} p(Y_i = 1, i \in \kappa, \text{and } Y_j = 0, j \in \Omega - \kappa | X)} \quad (9)$$

*and*

$$\exp(S^\omega) = \frac{p(Y_i = 1, i \in \omega, \text{and } Y_j = 0, j \in \Omega - \omega | X)}{P(Y_i = 0, i \in \Omega | X)} \quad (10)$$

The equivalence between graphical models and MVB model is given in the following theorem.

**Theorem 2.2** *Graphical model of general form (1) with 0/1 nodes is equivalent to multivariate Bernoulli model (2). And the followings are equivalent:*
  1. *There is no $|C|$-order interaction in $\{Y_i, i \in C\}$.*
  2. *There is no clique $C \in \Psi_K$ in the graph.*
  3. *$f^\omega = 0$ for all $\omega$ such that $C \subset \omega$.*

A proof is given in Appendix A. Theorem 2.2 states that there is a clique $C$ in the graphical model, if there is $\omega \supset C, f^\omega \neq 0$ in MVB model. The conditional independence specified in the graphical model can be fully formulated by MVB model.

**Example 2.1** *For a graph with $K$ nodes, the parameters in GM are $\{\Phi_\omega \mid \omega \in \Psi_K\}$, where $\Phi_\omega = \Phi_\Omega(Y_i = 1, i \in \omega, \text{and } Y_j = 0, j \in \Omega - \omega)$ is the potential function. We usually restrict $\Phi_\emptyset = 1$ to make the model identifiable. So there are $2^K - 1$ free parameters. Similarly, there are also $2^K - 1$ free parameters in MVB model $(f^1, \ldots, f^\Omega)$*

*When $K = 2$, $\Omega = \{1, 2\}, \mathcal{C} = \{\Omega\}$, define $\Phi_{11}$ to be the potential function $\Phi_\Omega(Y_1 = 1, Y_2 = 1; X)$ for simplicity, and define $\Phi_{10}, \Phi_{01}, \Phi_{00}$ similarly. The probability distribution with the GM parameterization is*

$$p(Y_1 = 1, Y_2 = 1|X) = \frac{1}{Z}\Phi_{11}, \quad p(Y_1 = 1, Y_2 = 0|X) = \frac{1}{Z}\Phi_{10},$$
$$p(Y_1 = 0, Y_2 = 1|X) = \frac{1}{Z}\Phi_{01}, \quad p(Y_1 = 0, Y_2 = 0|X) = \frac{1}{Z}\Phi_{00}$$

*The relation between GM and MVB model is*

$$f^1 = \log \frac{\Phi_{10}}{\Phi_{00}}, \quad f^2 = \log \frac{\Phi_{01}}{\Phi_{00}}, \quad f^{1,2} = \log \frac{\Phi_{11} \cdot \Phi_{00}}{\Phi_{01} \cdot \Phi_{10}}$$

*Note, the independence between $Y_1$ and $Y_2$ implies*

$$f^{1,2} = 0 \quad or \quad \log \frac{\Phi_{11} \cdot \Phi_{00}}{\Phi_{01} \cdot \Phi_{10}} = 0 \quad (11)$$

*Therefore, the sparseness in the conditional log odds ratios in MVB model gives a direct inference of the graph structure. But this property does not apply to GM.*

## 3. Structure Penalty

In many applications, the assumption of graphical models is that the graph structure has few large cliques. It is equivalent to the sparsity in higher order interactions in MVB model by Theorem 2.2. So we will impose a sparse penalty on the dependence structure to shrink higher order interactions.

Let $y(i) = (y_1(i), \ldots, y_K(i)), x(i) = (x_1(i), \ldots, x_p(i))$ be the $i$th data point. The augmented representation of the multivariate responses is:

$$\mathcal{Y}(i) = (y^1(i), \ldots, y^\omega(i), \ldots, y^\Omega(i)) \qquad (12)$$

There are $|\Psi_K| = 2^K - 1$ components in total[1]. Denote the number of components by $q$. In this paper, we consider the learning of the full model where $q = |\Psi_K|$. Suppose each function $f^\omega$ is in a Reproducing Kernel Hilbert Space (RKHS) $H^\omega$ with kernel $K^\omega$ (Wahba, 1990). The general penalized log likelihood model is:

$$\min I_\lambda(f) = L(f) + \lambda J(f) \qquad (13)$$

where the first term is the negative log-likelihood:

$$L(f) = \sum_{i=1}^{n} \Big( -\mathcal{Y}(i)^T f(x(i)) + b(f) \Big) \qquad (14)$$

and the second term $J(\cdot)$ is the structure penalty.

The objective is to obtain sparse estimation of the cliques by structure penalty on $f$. Consider the pairwise links. No link between $Y_s, Y_t$ in the graphical model means $f^\omega = 0$ for all $\omega \supset \{s, t\}$. For example, in Figure 1(b), $Y_1, Y_4$ are conditionally independent means $f^{1,4}, f^{1,2,4}, f^{1,3,4}, f^{1,2,3,4}$ are all zero. This objective is similar to sparse covariance matrix estimation in Gaussian data for neighborhood selection with lasso (Meinshausen and Buhlmann, 2006). However, our model will deal with higher order covariance structures that do not exist in Gaussian data. In addition, we not only consider the graph structure of responses $Y$ alone, but also the functions of predicative variables $X$ on $Y$.

To satisfy this intuition, the penalty is designed to shrink large cliques in the graph. Suppose in the true model, there is no interaction on clique $C$, then all $f^\omega$ should be zero, for $C \subset \omega$. The penalty is designed to shrink such $f^\omega$ to zero. The idea can be viewed as group lasso with overlaps. Group lasso (Yuan and Lin, 2006) leads to selection of variables in groups. It has consistent estimation when the groups are exclusive and union to the whole set. Jacob et al. (2009) considered the penalty on groups with arbitrary overlaps. Zhao et al. (2009) set up the general framework for hierarchical variable selections with overlapping groups, which we adopt here for the functions.

We consider the penalty guided by the structure in Figure 2. The guiding graph $T$ has $2^K - 1$ nodes: $1, \ldots, \omega, \ldots, \Omega$. With some abuse of notation, we use the element in $\Psi_k$ to index the node in $T$. There is an edge from $\omega_1$ to $\omega_2$ if and only if $\omega_1 \subset \omega_2$ and $|\omega_1| + 1 = |\omega_2|$. Domain knowledge can be applied here to design a different guiding structure. Jenatton et al. (2009) discussed how to define the groups to achieve different nonzero patterns.

---

1. In applications with large graphs, we only consider up to $m$'th interactions. We truncate higher order interactions and get $\sum_{k=1}^{m} \binom{K}{k}$ functions

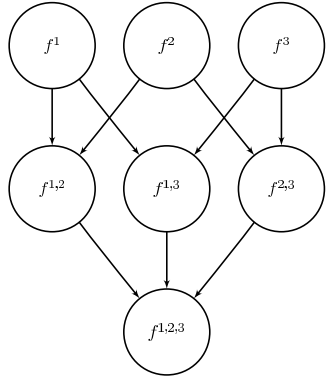

Figure 2: Hierarchical Guiding Structure for Penalty

Let $T_v = \{\omega \in \Psi_K | v \subset \omega\}$ be the subgraph rooted at $v$ in $T$, including all the descendants of $v$. Denote $f^{T_v} = (f^\omega)$, $\omega \in T_v$. All the functions are categorized into groups with overlaps as $G = (T_1, \ldots, T_\Omega)$. The structure penalty on the group $T_v$ of functions is:

$$J(f^{T_v}) = p_v \sqrt{\sum_{\omega \in T_v} \|f^\omega\|^2_{\mathcal{H}^\omega}} \tag{15}$$

where $p_v$ is the weight for the penalty on $T_v$ chosen empirically as $\frac{1}{|T_v|}$. Then the objective function is:

$$\min_f \quad I_\lambda(f) = L(f) + \lambda \sum_v p_v \sqrt{\sum_{\omega \in T_v} \|f^\omega\|^2_{\mathcal{H}^\omega}} \tag{16}$$

The following theorem shows that by solving the objective (16), $f^{\omega_1}$ will enter the model before $f^{\omega_2}$ if $\omega_1 \subset \omega_2$. That is to say, if $f^{\omega_1}$ does not exist, there will be no higher order interactions on $\omega_2$. The proof is given in Appendix A.

**Theorem 3.1** *Objective (16) is convex, thus the minimal is attainable. Let $\omega_1, \omega_2 \in \Psi_K$ and $\omega_1 \subset \omega_2$. If $f^*$ is the minimizer, that is $0 \in \partial I_\lambda(f^*)$ which is the subgradient of $I_\lambda$ at $f^*$, then $f^{*\omega_2} = 0$ almost sure if $f^{*\omega_1} = 0$.*

**Example 3.1** *If $K = 3$, $f = (f^1, f^2, f^3, f^{1,2}, f^{1,3}, f^{2,3}, f^{1,2,3})$. The grouped functions at node 1 in Figure 2 is $f^{T_1} = (f^1, f^{1,2}, f^{1,3}, f^{1,2,3})$. The objective is:*

$$\begin{aligned}
\min -l(y,f) + \lambda \Big( & p_1 \sqrt{\|f^1\|^2 + \|f^{1,2}\|^2 + \|f^{1,3}\|^2 + \|f^{1,2,3}\|^2} \\
& + p_2 \sqrt{\|f^2\|^2 + \|f^{1,2}\|^2 + \|f^{2,3}\|^2 + \|f^{1,2,3}\|^2} \\
& + p_3 \sqrt{\|f^3\|^2 + \|f^{1,3}\|^2 + \|f^{2,3}\|^2 + \|f^{1,2,3}\|^2} \\
& + p_4 \sqrt{\|f^{1,2}\|^2 + \|f^{1,2,3}\|^2} + p_5 \sqrt{\|f^{1,3}\|^2 + \|f^{1,2,3}\|^2} \\
& + p_6 \sqrt{\|f^{2,3}\|^2 + \|f^{1,2,3}\|^2} + p_7 \sqrt{\|f^{1,2,3}\|^2} \Big)
\end{aligned} \tag{17}$$

**Algorithm 1** Proximal Linearization Algorithm
---
**Input:** $c_0, \alpha_0, \zeta > 1, tol > 0$
**repeat**
  Choose $\alpha_k \in [\alpha_{min}, \alpha_{max}]$
  Solve Eq (19) for $d_k$
  **while** $\delta_k = I_\lambda(c_k) - I_\lambda(c_k + d_k) < \|d_k\|^3$ **do**
    // Insufficient decrease
    Set $\alpha_k = \max(\alpha_{min}, \zeta\alpha_k)$
    Solve Eq (19) for $d_k$
  **end while**
  Set $\alpha_{k+1} = \alpha_k/\zeta$
  Set $c_{k+1} = c_k + d_k$
**until** $\delta_k < tol$
---

## 4. Parameter Estimation

In this paper, we focus on the situation where the $\omega$th function space is $\mathcal{H}^\omega = \{1\} \oplus \mathcal{H}_1^\omega$. $\{1\}$ refers to the constant function space, and $\mathcal{H}_1^\omega$ is a RKHS with a linear kernel. The function $f^\omega \in \mathcal{H}^\omega$ has the form: $f^\omega(x) = c_0^\omega + \sum_{j=1}^p c_j^\omega x_j$. Its norm is $\|f^\omega\|_{\mathcal{H}^\omega}^2 = \|c^\omega\|^2$. Here, we denote $c^\omega = (c_0^\omega, \ldots, c_p^\omega)^T \in \mathbb{R}^{p+1}$ as a vector of length $p+1$ and $c = (c^\omega)_{\omega \in \Psi_K} \in \mathbb{R}^{\tilde{p}}$ be the concatenated vector of all parameters of length $\tilde{p} = (p+1) \cdot q$. Hence, the objective (16) is now:

$$\min_c \quad I_\lambda(c) = L(c) + \lambda \sum_v p_v \|c^{T_v}\| \tag{18}$$

where $c^{T_v} = (c^\omega)_{\omega \in T_v}$ is a $(p+1) \cdot |T^v|$ vector.

To solve (18), we iteratively solve the following proximal linearization problem (Wright, 2010):

$$\min_c L_k + \nabla L_k^T(c - c_k) + \frac{\alpha_k}{2}\|c - c_k\|^2 + \lambda J(c) \tag{19}$$

where $L_k = L(c_k)$, $\alpha_k$ is a positive scalar chosen adaptively. With slight abuse of notation, we use $c_k$ to denote the vector of all parameters at $k$th step. Algorithm 1 summarized the framework of solving (18). Following the analysis in Wright (2010), we can show that the proximal linearization algorithm will converge for negative log-likelihood loss function plus group lasso type penalties with overlaps.

However, solving group lasso with overlaps is not trivial due to the non-smoothness at the singular point. In recent years, several papers have addressed this problem. Jacob et al. (2009) duplicated the variables in the design matrix that appear in group overlaps, then solved the problem as group lasso without overlaps. Kim and Xing (2010) reparameterized the group norm with additional dummy variables. They alternatively optimized for the model parameters and the dummy variables at each iteration. The method performs efficiently on quadratic loss function for Gaussian data. But optimizing alternatively over two sets of parameters might not scale well on penalized logistic regression.

In this paper, we solve (19) by its smooth and convex dual problem proposed by Liu and Ye (2010). Let $Z = \{v \in \Psi_K | \|c^{T_v}\| = 0\}$, and $\bar{Z} = \Psi_K - Z$ be the complement. Define $s_v$, $v \in \Psi_K$ as:

$$s_v \in \mathbb{S}_v = \{s = (s^\omega)_{\omega \in \Psi_K} \mid s \in \mathbb{R}^{\tilde{p}}, \|s\| \leq \lambda p_v, s^\omega = 0 \text{ if } \omega \in T_v\} \quad (20)$$

then the subgradient of (19) is:

$$\nabla L + \alpha_k(c - c_k) + \sum_{v \in Z} s_v + \sum_{u \in \bar{Z}} r_u \quad (21)$$

where $s_v$ is the subgradient of $\lambda p_v \|c^{T_v}\|$ for $v \in Z$ and $r_u$ is the subgradient for $u \in \bar{Z}$:

$$r_u = \arg\max_{s_u} \langle s_u, c \rangle, \text{ for } u \in \bar{Z} \quad (22)$$

The subgradient $s_v$ is in a unit ball of certain subspace of $\mathbb{R}^{\tilde{p}}$. These subspaces are not perpendicular to each other. Thus, $s_v$'s are not separable, and closed form solution of (19) cannot be obtained. We solve the proximal subproblem (19) by its smoothing and convex dual problem. Note (19) is equivalent to:

$$\min_{c \in \mathbb{R}^{\tilde{p}}} \max_{S \in \mathbb{S}} \phi(c, S) = \quad (23)$$
$$\nabla L_k^T(c - c_k) + \frac{\alpha_k}{2}\|c - c_k\|^2 + \sum_{v \in \Omega} \langle s_v, c \rangle$$

where $S$ is a $\tilde{p} \times q$ matrix whose columns are $s_v$. $\mathbb{S} = \{S | S = (s_1, \ldots, s_v, \ldots, s_\Omega), s_v \in \mathbb{S}_v \text{ for } v \in \Psi_K\}$ is the feasible region of $S$. Since $\phi(\cdot, S)$ is lower semicontinuous and $\phi(c, \cdot)$ is upper semicontinuous, there exists a saddle point and the max and min are exchangeable. The solution of minimizing $\phi(c, S)$ is:

$$\tilde{c} = \arg\min_c \phi(c, S) = c_k - \frac{1}{\alpha_k} \nabla L_k - \frac{1}{\alpha_k} \sum_v s_v \quad (24)$$

Substitute $\tilde{c}$ back into (23), we have the dual problem of (19) as:

$$\max_{S \in \mathbb{S}} \eta(S) = -\frac{1}{2}\|\sum_v s_v\|^2 + (\alpha_k c_k - \nabla L_k)^T \sum_v s_v \quad (25)$$

Following the proof in Liu and Ye (2010), we can show that $\eta(S)$ is convex and Lipschitz continuous. The differential is $\alpha_k \tilde{c} e^T$ where $e \in \mathbb{R}^{\tilde{p}}$ is a vector of ones. Hence, (25) can be solved by existing gradient methods. We use the accelerated gradient descent method implemented in (Liu et al., 2009).

## 5. Experiments

### 5.1 Simulation Study

The simulations are performed to evaluate the learning accuracy of our method. The graphs of the response variables are depicted in Figure 1. In the simulation, we assume the most general distribution family for each graphical model according to its graph structure. For example, in Model 1, we have 4 response variables, $(Y_1, Y_2, Y_3, Y_4)$. There is a triangular clique $(Y_1, Y_2, Y_3)$, but $Y_4$ is independent with the other response variables. In this case, we have a 15 conditional log odds ratios to estimate. In the true model, the non-zero functions are $\{f^1, f^2, f^3, f^{1,2}, f^{1,3}, f^{2,3}, f^{1,2,3}, f^4\}$. In Model 3, there are 255 functions and 25 of them are nonzero. In Model 4, there are 1023 functions and 25 of them are nonzero.

The predictive variables $X = (X_1, \ldots, X_5)$ are independently generated from multivariate Gaussian distribution with mean 0 and variance 1. Each $f^\omega$ has 6 parameters to estimate (1 of them is the intercept). These parameters, $c_j^\omega, j = 1, \cdots, 5$, are uniformly sampled from $\{-5, -4, \cdots, 5\}$. We set the intercepts $c_0^\omega$ in main effects to 1, those in second or higher order interactions to 2. Each $Y$ is randomly selected proportional to the probability in equation (2), where $f = Xc$. We generate 100 datasets for each graph structure in Figure 1 to evaluate the learning accuracy. The sample size in each dataset is 1000.

The tuning parameter $\lambda$ is chosen by two different tuning methods: 1) GACV (generalized approximation cross validation), 2) BGACV (B-type GACV). The details of the tuning methods are discussed in Appendix B.

Table 2: # of True Positive and False Positive Functions

|  | $f^{1,2}$ | $f^{1,3}$ | $f^{2,3}$ | $f^{3,4}$ | $f^{1,2,3}$ | $f^{5,7,8}$ | $f^{5,6,7,8}$ | FP |
|---|---|---|---|---|---|---|---|---|
|  | Model 1 | | | | | | | |
| GACV | 100 | 100 | 100 | 99 | 89 | - | - | 123 |
| BGACV | 83 | 89 | 86 | 68 | 17 | - | - | 5 |
|  | Model 2 | | | | | | | |
| GACV | 100 | 99 | 100 | 99 | 83 | - | - | 321 |
| BGACV | 92 | 93 | 92 | 83 | 34 | - | - | 46 |
|  | Model 3 | | | | | | | |
| GACV | 95 | 89 | 79 | 98 | 57 | 77 | 31 | 394 |
| BGACV | 37 | 18 | 21 | 92 | 0 | 36 | 0 | 143 |
|  | Model 4 | | | | | | | |
| GACV | 91 | 98 | 96 | 88 | 59 | 49 | - | 654 |
| BGACV | 70 | 72 | 69 | 66 | 0 | 0 | - | 134 |

In Table 2, we count, for each function $f^\omega$, the number of runs out of the 100 replications in which $f^\omega$ is recovered ($\|c^\omega\| \neq 0$). The recovered functions in the true model are considered as true positive; while the others not in the true model are false positive. Since the main effects are always detected correctly, they are not listed in the table. The structure penalty is efficient in recognizing strong interactions in the responses, such as the interaction between $Y_1, Y_2$. But its performance on higher order interactions will be affected in more complex graph structures, e.g. $f^{1,2,3}$ in Model 3 and 4.

Compared to GACV, BGACV tends to achieve more sparse results in general, because they have large penalties on the degrees of freedom of the estimated model. On the contrary, GACV will discover more true positive functions with the cost of higher false positive rate.

In Figure 3, we show the learning results in terms of true positive rate (TPR) with increasing sample size from 100 to 1000. The experimental setting is the same as before.

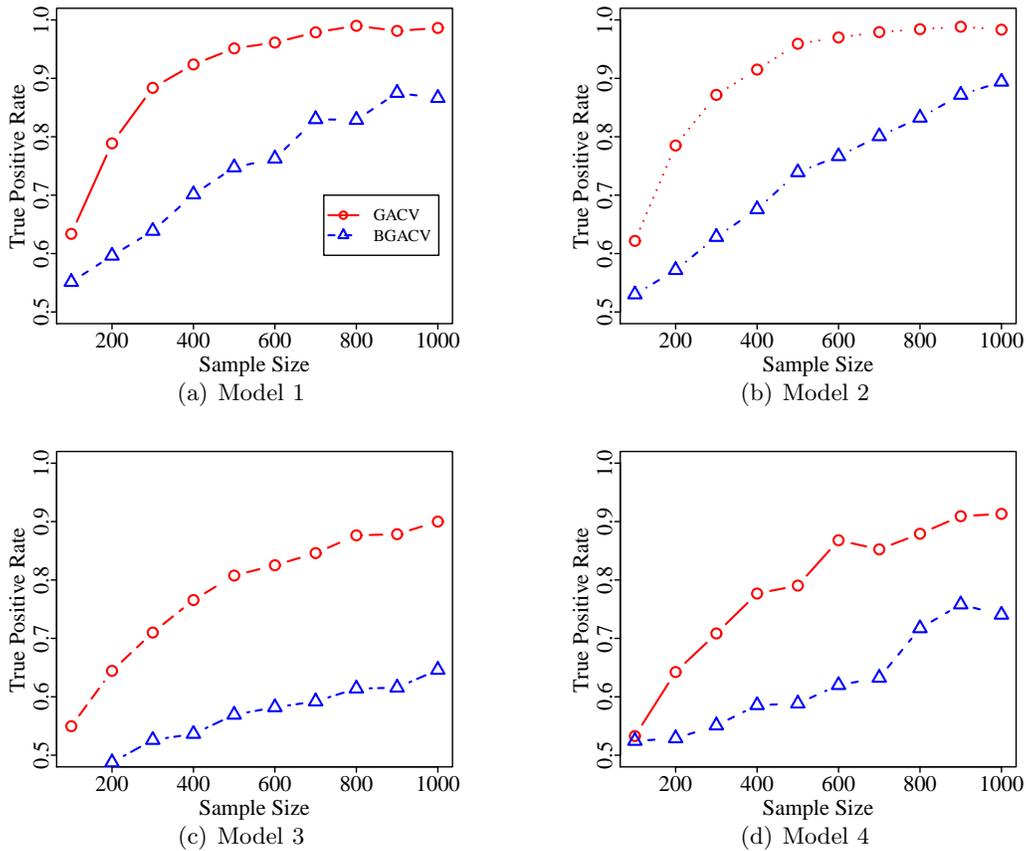

Figure 3: True Positive Rate (TPR) of Graph structure learning with increasing sample size from 100 to 1000 by different tuning methods: GACV, BGACV. The TPR curves (GACV) have converged for M1, M2 after 500 samples. But for M3 and M4, larger sample size is required to achieve convergence, since there are 1023 functions to estimate. BGACV is more conservative in selecting non-zero functions.

The TPRs are improving along with the increasing sample size. Compared to Model 1 and 2, the algorithm needs substantially larger sample size to achieve high TPR on Model 3 and 4. GACV achieves better true positive rate in all four graphical models. This tuning method will obtain less sparse models compared to BGACV.

### 5.2 Census Bureau County Data

We use the county data from U.S. Census Bureau[2] to validate our method. It includes demographic data for all counties in United States, covering population, employment, votes (Scammon et al., 2005), and etc. We delete the counties which have missing value in the columns we are interested, and get 2668 data entries in total. The outcomes for this study are summarized in Table 3. "Vote" is coded as 1 if the Republican candidate won in the 2004 presidential election. To dichotomize the rest response variables, the national mean is selected as a threshold. The third column in the table gives the percentage of positive in the data. The features in the model are: percentage of housing unit change; government expenditure; population percentage of 3 ethnic groups (White, African, Asian), people foreign born, people over 65, people under 18, people with high school education, and people with bachelor degree; birth rate; death rate.

---
2. http://www.census.gov/statab/www/ccdb.html

Table 3: Selected Response variables

| Response | Description | Positive% |
|---|---|---|
| Vote | 2004 votes for Republican presidential candidate | 81.11 |
| Poverty | Persons in poverty | 52.70 |
| VCrime | Violent Crime, eg. murder, robbery | 23.09 |
| PCrime | Property Crime, eg. burglary | 6.82 |
| URate | Unemployment Rate | 51.35 |
| PChange | Population change in percentage from 2000 to 2006 | 64.96 |

In the experiment, We first standardize the data to be mean 0 variance 1. Then, we can get an estimated graphical model with every fixed $\lambda$. Adjusting the regularization parameter $\lambda$ from 0.3289 to 0.1389, we will discover new interactions entering the model. The graph structure of $\lambda = 0.1559$ (chosen by cross validation) is shown in Figure 4. The first number on edge indicates the order of the link entering the model, while the second one is the corresponding $\lambda$. The unemployment rate plays an important role as a hub in the graph. It is strongly related to poverty and crimes. Population change is negatively related to violent crimes, as well as to unemployment rate.

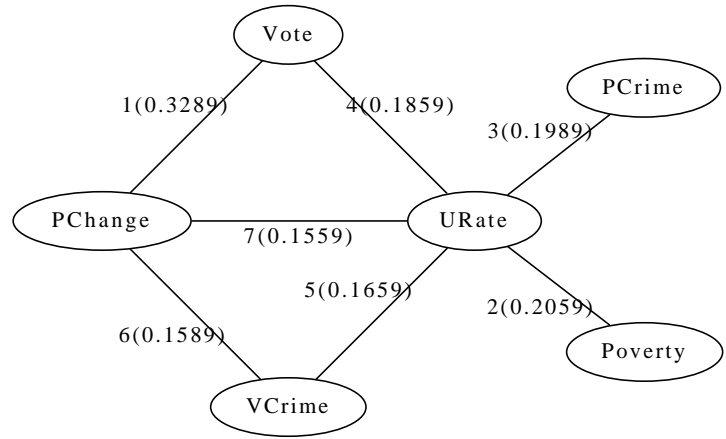

Figure 4: Correlation of Response in the county data from Census Bureau. The first number on the edge shows the order at which the link is recovered. The number in bracket denotes the corresponding $\lambda$.

We analyze the link between "Vote" and "PChange", which is recovered by our method before any other links. The marginal correlation between them (without conditioning on $X$) is 0.0389, which is the second smallest absolute value in the correlation matrix. The partial correlation method (Peng et al., 2009) is taken as an example to show how the links of response variables are discovered without considering $X$. The link between "Vote" and "PChange" is the tenth recovered link using R package *space*. The reason is that after taking features into account, the dependence structure of response variables may change. The main contribution in this case is "percentage of housing unit change" ($X_1$) and "population percentage of people over 65" ($X_2$). Part of the fitted model is shown below:

$$f^{Vote} = 0.0463 \cdot X_1 + 0.0877 \cdot X_2 + \cdots$$
$$f^{PChange} = 0.2315 \cdot X_1 - 0.0942 \cdot X_2 + \cdots$$
$$f^{Vote, PChange} = 0.0211 \cdot X_1 - 0.0115 \cdot X_2 + \cdots$$

The main effects suggest that with increase of housing units, the counties tend to increase in population and vote for Republican. With increase of people over 65, the counties tend to lose population, but still more likely to vote for Republican. The interaction function reveals that as housing units increase, the counties are more likely to have both positive results for "Vote" and "PChange". But this tendency will be counteracted by the increase of people over 65: the responses are less likely to take both positive values.

## 6. Conclusions and Future Work

The structure penalty on the multivariate Bernoulli model can efficiently learn the graph structure, which indicates the conditional independence of the response variables. In this paper, we only consider linear models for the conditional log odds ratios. It can be extended to smoothing spline ANOVA model with more freedom by choosing the kernels. And Theorem 3.1 holds natually. It is also interesting to see how the penalty will improve the prediction power compared to large margin methods.

Another extension will be the selection of features for each $f^\omega$. The sparsity in each function requires the sparsity within each group. But the graph structure would change with different selected features. One application is to discover the relations of multiple symptoms or clinical responses and how they are effected by the environmental and genetical covariates. Smoking could be significant for many diseases and their interactions, but other covariates, such as taking Vitamin might be only related to a subset of the symptoms. As a result, we will investigate sparse penalties within each function.

## Acknowledgments

Research is supported in part by NIH Grant EY09946, NSF Grant DMS-0906818 and ONR Grant N0014-09-1-0655.

# References

O. Banerjee, L. El Ghaoui, and A. d'Aspremont. Model selection through sparse maximum likelihood estimation for multivariate gaussian or binary data. *The Journal of Machine Learning Research*, 9:485–516, 2008. ISSN 1532-4435.

Z. Duan, L. Lu, and C. Zhang. Collective annotation of music from multiple semantic sategories. In *Proceedings of 9th International Conference on Music Information Retrieval*, pages 237–242, Philadelphia, USA, 2008.

F. Gao, G. Wahba, R. Klein, and B. Klein. Smoothing Spline ANOVA for multivariate Bernoulli observations, with application to ophthalmology data. *Journal of the American Statistical Association*, 96(453):127, 2001.

J. Honorio and D. Samaras. Multi-Task Learning of Gaussian Graphical Models. In *Proceedings of the 27th International Conference on Machine learning*, pages 447–454, Haifa, Israel, 2010.

L. Jacob, G. Obozinski, and J.P. Vert. Group Lasso with overlap and graph Lasso. In *Proceedings of the 26th Annual International Conference on Machine Learning*, pages 433–440, 2009.

R. Jenatton, J.Y. Audibert, and F. Bach. Structured variable selection with sparsity-inducing norms. *arXiv:0904.3523*, 2009.

S. Kim and E.P. Xing. Tree-guided group lasso for multi-task regression with structured sparsity. In *Proceedings of 27th International Conference on Machine Learning*, pages 543–550, Haifa, Israel, 2010.

J. Lafferty, A. McCallum, and F. Pereira. Conditional random fields: Probabilistic models for segmenting and labeling sequence data. In *Proceedings of the 18th International Conference on Machine Learning*, pages 282–289, 2001.

J. Liu and J. Ye. Fast overlapping group lasso. *arXiv:1009.0306v1*, 2010.

J. Liu, S. Ji, and J. Ye. *SLEP: Sparse Learning with Efficient Projections*. Arizona State University, 2009. URL http://www.public.asu.edu/~jye02/Software/SLEP.

Xiwen Ma. *Penalized Regression in Reproducing Kernel Hilbert Spaces With Randomized Covariate Data*. PhD thesis, Department of Statistics, University of Wisconsin-Madison, 2010.

N. Meinshausen and P. Buhlmann. High-dimensional graphs and variable selection with the lasso. *The Annals of Statistics*, 34(3):1436–1462, 2006.

J. Peng, P. Wang, N. Zhou, and J. Zhu. Partial correlation estimation by joint sparse regression models. *Journal of the American Statistical Association*, 104(486):735–746, 2009.

P. Ravikumar, M.J. Wainwright, and J. Lafferty. High-dimensional Ising model selection using l1-regularized logistic regression. *Annals of Statistics*, 38(3):1287–1319, 2010.

R.M. Scammon, A.V. McGillivray, and R. Cook. *America Votes 26: 2003-2004, Election Returns By State*. CQ Press, 2005. ISBN 9781568029740.

W. Shi, G. Wahba, S. Wright, K. Lee, R. Klein, and B. Klein. LASSO-Patternsearch algorithm with application to ophthalmology and genomic data. *Statistics and its Interface*, 1(1):137, 2008.

TP Speed and HT Kiiveri. Gaussian Markov distributions over finite graphs. *Annals of Statistics*, 14(1):138–150, 1986. ISSN 0090-5364.

C. Sutton, A. McCallum, and K. Rohanimanesh. Dynamic conditional random fields: Factorized probabilistic models for labeling and segmenting sequence data. *The Journal of Machine Learning Research*, 8:693–723, 2007. ISSN 1532-4435.

R. Szeliski, R. Zabih, D. Scharstein, O. Veksler, V. Kolmogorov, A. Agarwala, M. Tappen, and C. Rother. A comparative study of energy minimization methods for markov random fields with smoothness-based priors. *IEEE Transactions on Pattern Analysis and Machine Intelligence*, 30:1068–1080, June 2007. ISSN 0162-8828.

G. Wahba. *Spline Models for Observational Data*. Society for Industrial Mathematics, 1990.

M.J. Wainwright and M.I. Jordan. Graphical models, exponential families, and variational inference. *Foundations and Trends® in Machine Learning*, 1:1–305, 2008. ISSN 1935-8237.

O. Williams, A. Blake, and R. Cipolla. The Variational Ising Classifier (VIC) algorithm for coherently contaminated data. *Advances in Neural Information Processing Systems*, 17:1497–1504, 2004.

S.J. Wright. Accelerated block-coordinate relaxation for regularized optimization. Technical report, Department of Computer Science, University of Wisconsin-Madison, 2010.

D. Xiang and G. Wahba. A generalized approximate cross validation for smoothing splines with non-Gaussian data. *Statistica Sinica*, 6:675–692, 1996. ISSN 1017-0405.

M. Yuan and Y. Lin. Model selection and estimation in regression with grouped variables. *Journal of the Royal Statistical Society: Series B (Statistical Methodology)*, 68(1):49–67, 2006.

P. Zhao, G. Rocha, and B. Yu. The composite absolute penalties family for grouped and hierarchical variable selection. *Annals of Statistics*, 37(6A):3468–3497, 2009.

## Appendix A. Proof

### A.1 Proof of Theorem 2.2

**Proof** Given graphical model (1), let $y_C^\omega$ be a realization of $y_C$ such that $y_C^\omega = \{y_i^\omega \mid i \in C\}$ where $y_i^\omega = 1$ if $i \in \omega$ and $y_i^\omega = 0$ otherwise. Let the odd-even partition of the power set of $\omega$ defined as in Lemma 2.1. The conditional log odds ratios in MVB model are:

$$f^\omega(x) = \log \frac{\prod_{\kappa \in \Psi_{even}^\omega} \prod_{C \in \mathcal{C}} \Phi_C(y_C^\kappa; x)}{\prod_{\kappa \in \Psi_{odd}^\omega} \prod_{C \in \mathcal{C}} \Phi_C(y_C^\kappa; x)} \quad (26)$$

$$b(f) = \log \frac{Z(x)}{\prod_{C \in \mathcal{C}} \Phi_C(0; x)}$$

Conversely, given the MVB model of 2, the cliques can be determined by the nonzero $f^\omega$: clique $C$ exists if $C = \omega$ and $f^\omega \neq 0$. Then the maximal cliques can be inferred from the graph structure. And suppose they are $C_1, \ldots, C_m$. Let $\omega_i$ be $\psi(\omega_i) = C_i, i = 1, \ldots, m$, and $\kappa_1 = \emptyset$, $\kappa_i$ be $\psi(\kappa_i) = C_i \cap (C_{i-1} \cup \cdots \cup C_1), i = 2, \ldots, m$. Then one possible parameterization is:

$$\Phi_{C_i}(y_{C_i}; x) = \exp\left(S^{\omega_i}(y; x) - S^{\kappa_i}(y; x)\right) \quad (27)$$

$$Z(x) = \exp(b(f))$$

Therefore, graphical model (1) with bivariate outcomes is equivalent to the MVB Model (2).

In the latter part of the theorem, $1 \Rightarrow 2$ and $3 \Rightarrow 1$ follows naturally from the Markov property of graphical models. To show $2 \Rightarrow 3$, notice that whenever $\kappa \cap C = \kappa' \cap C$, $y_C^\kappa = y_C^{\kappa'}$. For any possible $v = \kappa \cap C$, $\kappa' \in \{\kappa | \kappa = v \cup u, \text{ s.t. } u \subset \omega - v\}$ will give $\kappa' \cap C = v$. There are $2^{|\omega - v|}$ such $\kappa'$ in total due to the choice of $u$. Also, they appear in the nominator and denominator of equation (26) equally. So, for any $C \in \mathcal{C}$,

$$\prod_{\kappa \in \Psi_{even}^\omega} \Phi_C(y_C^\kappa; x) = \prod_{\kappa \in \Psi_{odd}^\omega} \Phi_C(y_C^\kappa; x) \quad (28)$$

It follows that $f^\omega = 0$ by (26). ∎

### A.2 Proof of Theorem 3.1

**Proof** We give the proof for the linear case. The convexity of $I_\lambda$ is easy to check, since $L$ and $J(f^{T_v})$ are all convex in $c$. Suppose there is some $\omega_2 \supset \omega_1$ s.t. $c^{*\omega_2} \neq 0$, by the groups constructed through Figure 2, $\|c^{*T_v}\| \neq 0$ for all $v \subset \omega_1$. So the partial derivative of objective (18) with respect to $c^{\omega_1}$ is

$$\frac{\partial L}{\partial c^{\omega_1}} + \lambda \sum_{v \subset \omega_1} p_v \frac{c^{*\omega_1}}{\|c^{*T_v}\|} = 0 \quad (29)$$

Hence, the probability of $\{\exists \omega_2 \supset \omega_1 \text{ s.t } c^{*\omega_2} \neq 0\}$ equals to the probability that $\frac{\partial L}{\partial c^{\omega_1}} = 0$, which is 0. ∎

## Appendix B. Tuning

For $i$-th data, we have:

$$S_i^\omega = S^\omega(x(i)) \tag{30}$$

$$b_i = b(f(x(i))) = \log\left(1 + \sum_\omega \exp S_i^\omega\right) \tag{31}$$

Then, the mean of the augmented response $\mathcal{Y}(i)$ in the multivariate Bernoulli model is:

$$\mu(i) = E[\mathcal{Y}(i)|x(i), f] = (\mu^1(i), \cdots, \mu^\kappa(i), \cdots, \mu^\Omega(i)) \tag{32}$$

$$\text{where } \mu^\kappa(i) = \frac{\partial b_i}{\partial f^\kappa} = \frac{\sum_{\omega \in T_\kappa} \exp S_i^\omega}{\exp b_i} \tag{33}$$

The $q \times q$ covariance matrix of the augmented response is:

$$W(i) = var(\mathcal{Y}(i)|x(i), f) \tag{34}$$

where the $(\alpha, \beta)$-th element of $W(i)$ is:

$$W_{\alpha,\beta}(i) = \frac{\partial^2 b_i}{\partial f^\alpha (\partial f^\beta)^T} = \frac{\sum_{\omega \in T_\alpha \cap T_\beta} \exp S_i^\omega}{\exp b_i} - \mu^\alpha(i) \cdot \mu^\beta(i) \tag{35}$$

Let $R_v$ be a $\tilde{p} \times \tilde{p}$ diagonal matrix whose $(i, i)$-th element is 1 if $c_i \neq 0$. Then, the $v$-th group penalty in (18) can be written as:

$$J(f^{T_v}) = p_v \sqrt{\sum_{\omega \in T_v} \|f^\omega\|^2} = p_v \|R_v c\| \tag{36}$$

Note $R_v$ is symmetric and $R_v \cdot R_v = R_v$, direct calculation yields the derivative and Hessian of the penalty term:

$$\frac{\partial J}{\partial c} = \sum_{v: R_v c \neq 0} p_v \frac{R_v c}{\|R_v c\|} \tag{37}$$

$$\frac{\partial^2 J}{\partial c \partial c^T} = \sum_{v: R_v c \neq 0} p_v J_v = \sum_{v: R_v c \neq 0} p_v \frac{R_v(\|R_v c\|^2 I - c \cdot c^T) R_v}{\|R_v c\|^3} \tag{38}$$

where $J_v \doteq (R_v(\|R_v c\|^2 I - c \cdot c^T) R_v)/\|R_v c\|^3$. Denote the grand design matrix as:

$$D = \begin{pmatrix} D(1)^T & \cdots & D(n)^T \end{pmatrix}^T \tag{39}$$

$$\text{where } D(i) = \begin{pmatrix} x(i)^T & 0 & \cdots & 0 \\ 0 & x(i)^T & \cdots & 0 \\ \vdots & \vdots & \ddots & \vdots \\ 0 & 0 & \cdots & x(i)^T \end{pmatrix} \tag{40}$$

Suppose there are $N$ non-zero elements of $c$ at location $\{a_1, \ldots, a_N\}$. Let $\tilde{D}$ be the matrix composed by the $a_1, \ldots, a_N$th column of $D$. Then, the Hessian matrix of $I_\lambda$ in (16) is:

$$\frac{\partial^2 I_\lambda}{\partial c \partial c^T} = \frac{\partial^2 L}{\partial c \partial c^T} + \lambda \frac{\partial^2 J}{\partial c \partial c^T} = \tilde{D}^T W \tilde{D} + \lambda \sum_{v: R_v c \neq 0} p_v J_v \tag{41}$$

$$\tag{42}$$

Let $H$ be the $nq \times nq$ influence matrix that implies

$$f_{\lambda,\epsilon} - f_\lambda \approx H \epsilon \tag{43}$$

where $\epsilon$ is a small perturbation on $\mathcal{Y}$; $f_\lambda = D c_\lambda$ is the estimated function value with tuning parameter $\lambda$; and $f_{\lambda,\epsilon}$ is the estimated function value with the perturbation. Then, the analysis of the first order Taylor expansion of $\frac{\partial I_\lambda}{\partial c}(\mathcal{Y} + \epsilon, c_{\lambda,\epsilon})$ leads to the formulation of $H$ as follows (refer to (Xiang and Wahba, 1996) and (Ma, 2010) Chapter 3 for more details)

$$H = \tilde{D} \Big( \frac{\partial^2 I_\lambda}{\partial c \partial c^T} \Big)^{-1} \tilde{D}^T = \tilde{D} \Big( \tilde{D}^T W \tilde{D} + \lambda \sum_{v: R_v c \neq 0} p_v J_v \Big)^{-1} \tilde{D}^T \tag{44}$$

The $(i,j)$-th $q \times q$ submatrix of $H$ is

$$H(i,j) = \tilde{D}(i)^T \Big( \frac{\partial^2 I_\lambda}{\partial c \partial c^T} \Big)^{-1} \tilde{D}(j) \tag{45}$$

Let $Q(i) = I - H(i,i)W(i)$ for $i = 1, \ldots, n$, define the generalized average matrix, denoted as $\bar{Q}$, of $\{Q(i), i = 1, \ldots, n\}$ as follows

$$\bar{Q} = (\delta - \gamma) I_{q \times q} + \gamma \cdot e e^T = \begin{pmatrix} \delta & \gamma & \cdots & \gamma \\ \gamma & \delta & \cdots & \gamma \\ \vdots & \vdots & \ddots & \vdots \\ \gamma & \gamma & \cdots & \delta \end{pmatrix} \tag{46}$$

where $e$ is the unit vector of length $q$ and

$$\delta = \frac{1}{nq \sum_{i=1}^n tr(Q(i))}, \quad \gamma = \frac{1}{nq(q-1)} \big[ e^T Q(i) e - tr(Q(i)) \big] \tag{47}$$

Let $\bar{H}$ be the generalized average of $\{H(i,i), i = 1, \cdots, n\}$, the GACV score is

$$GACV(\lambda) = OBS(\lambda) + \frac{1}{n} \sum_{i=1}^n \mathcal{Y}(i)^T \bar{Q}^{-1} \bar{H} \big( \mathcal{Y}(i) - \mu(i) \big) \tag{48}$$

where

$$OBS(\lambda) = \frac{1}{n} \Big[ -\mathcal{Y}(i)^T f_\lambda(x(i)) + b(f_\lambda(x(i))) \Big] \tag{49}$$

is the observed log-likelihood.

The degrees of freedom of multivariate Bernoulli data is generally difficult to obtain. But we can have a good approximation from GACV (Shi et al., 2008) as

$$\hat{df}(\lambda) = \sum_{i=1}^{n} \mathcal{Y}(i)^T \bar{Q}^{-1} \bar{H} \big( \mathcal{Y}(i) - \mu(i) \big) \tag{50}$$

So the BGACV score can be defined as

$$BGACV(\lambda) = OBS(\lambda) + \frac{1}{n} \frac{\log n}{2} \sum_{i=1}^{n} \mathcal{Y}(i)^T \bar{Q}^{-1} \bar{H} \big( \mathcal{Y}(i) - \mu(i) \big) \tag{51}$$